\documentclass[a4paper]{article}

\usepackage[english]{babel}
\usepackage[utf8x]{inputenc}
\usepackage[T1]{fontenc}

\usepackage[a4paper,top=3cm,bottom=2cm,left=3cm,right=3cm,marginparwidth=1.75cm]{geometry}

\usepackage{cite}
\usepackage{amsmath}
\usepackage{graphicx}
\usepackage{subcaption}
\usepackage[export]{adjustbox}

\usepackage{acronym}

\title{ADaPTION: Toolbox and Benchmark for Training Convolutional Neural Networks with Reduced Numerical Precision Weights and Activation}
\author{Moritz B. Milde, Daniel Neil, Alessandro Aimar,\\ Tobi Delbruck and Giacomo Indiveri}

\begin{document}
\acrodef{cnn}[CNN]{Convolutional Neural Network}
\acrodef{dnn}[DNN]{Deep Neural Network}
\acrodef{relu}[ReLU]{Rectified Linear Unit}
\maketitle

\begin{abstract}
\acp{dnn} and \acp{cnn} are useful for many practical tasks in machine learning.
Synaptic weights, as well as neuron activation functions within the deep network are typically stored with high-precision formats, e.g. 32 bit floating point. 
However, since storage capacity is limited and each memory access consumes power, both storage capacity and memory access are two crucial factors in these networks.
Here we present a method and present the ADaPTION toolbox to extend the popular deep learning library Caffe to support training of deep \acp{cnn} with reduced numerical precision of weights and activations using fixed-point notation. 
ADaPTION includes tools to measure the dynamic range of weights and activations.
Using the ADaPTION tools, we quantized several \acp{cnn} including VGG16 down to 16-bit weights and activations with only 0.8\% drop in Top-1 accuracy.
The quantization, especially of the activations, 
leads to increase of up to 50\% of sparsity especially in early and intermediate layers, which we exploit to skip multiplications with zero, thus performing faster and computationally cheaper inference.
\end{abstract}
\begin{figure}[!h]
\centering
\includegraphics[width=0.8\textwidth]{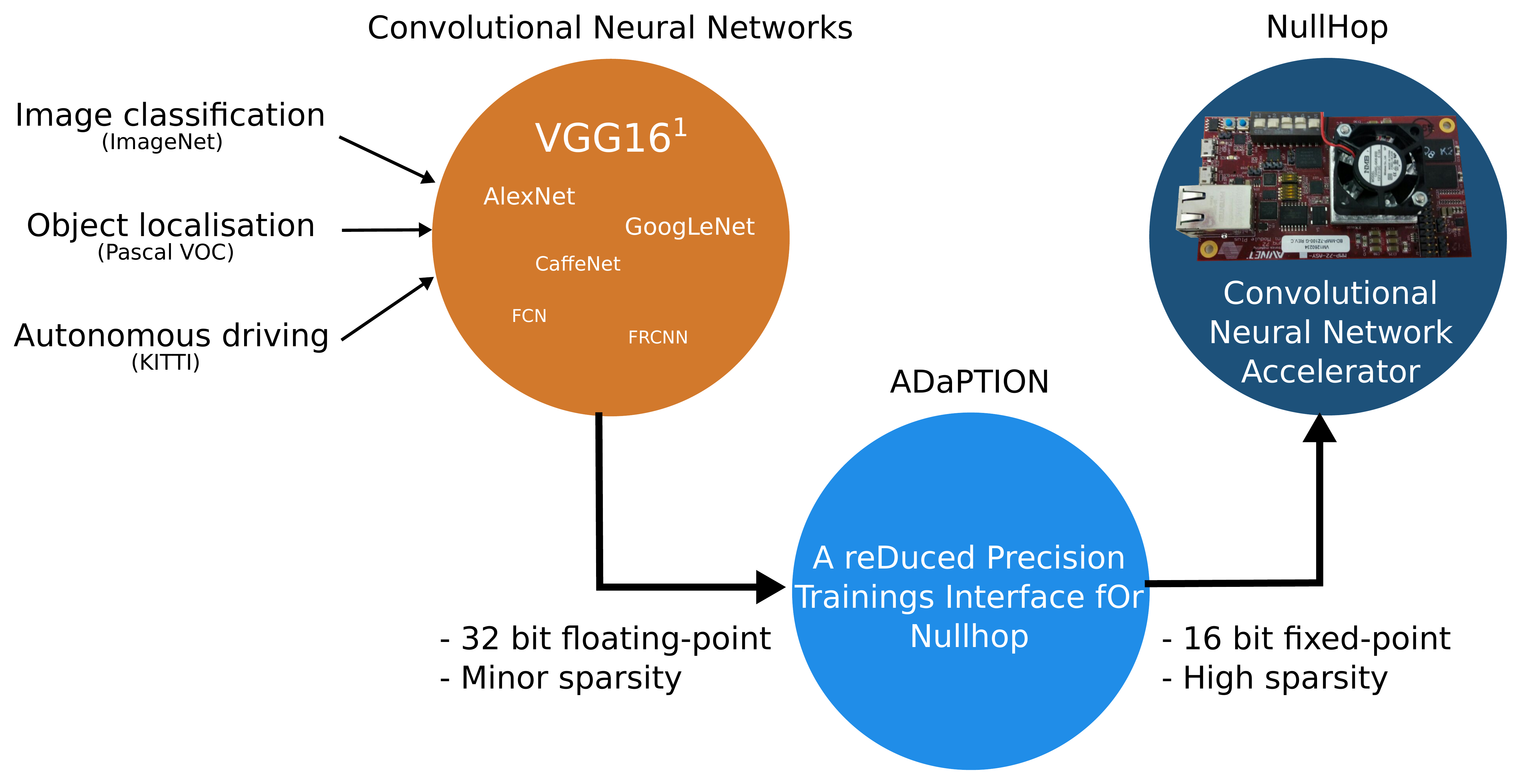}
\caption{ADaPTION allows the adaptation of high-precision \ac{cnn} models to a reduced precision fixed-point representation. This adaption of weight and activation precision allows the execution of these networks on fixed-point \ac{cnn} hardware accelerators, such as Nullhop \cite{Aimar_etal17}.}
\label{fig:motivation}
\end{figure}
\section{Introduction}
\begin{figure*}[!t]
    \centering
    \begin{subfigure}[b]{0.45\textwidth}
        \centering
        \includegraphics[width=\textwidth, valign=t]{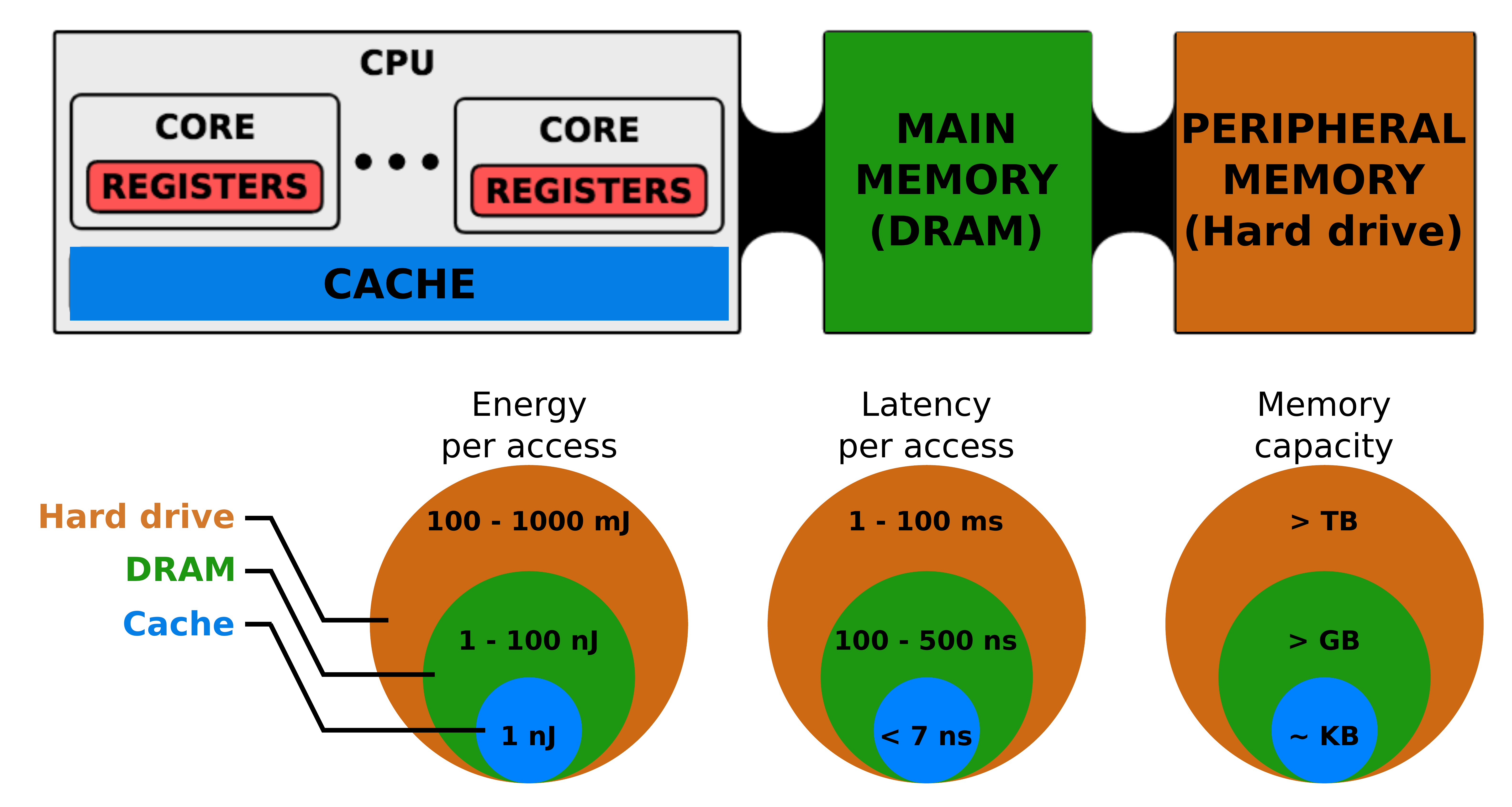}
        \caption{}
    \end{subfigure}%
    \begin{subfigure}[b]{0.45\textwidth}
        \centering
        \includegraphics[width=\textwidth, valign=t]{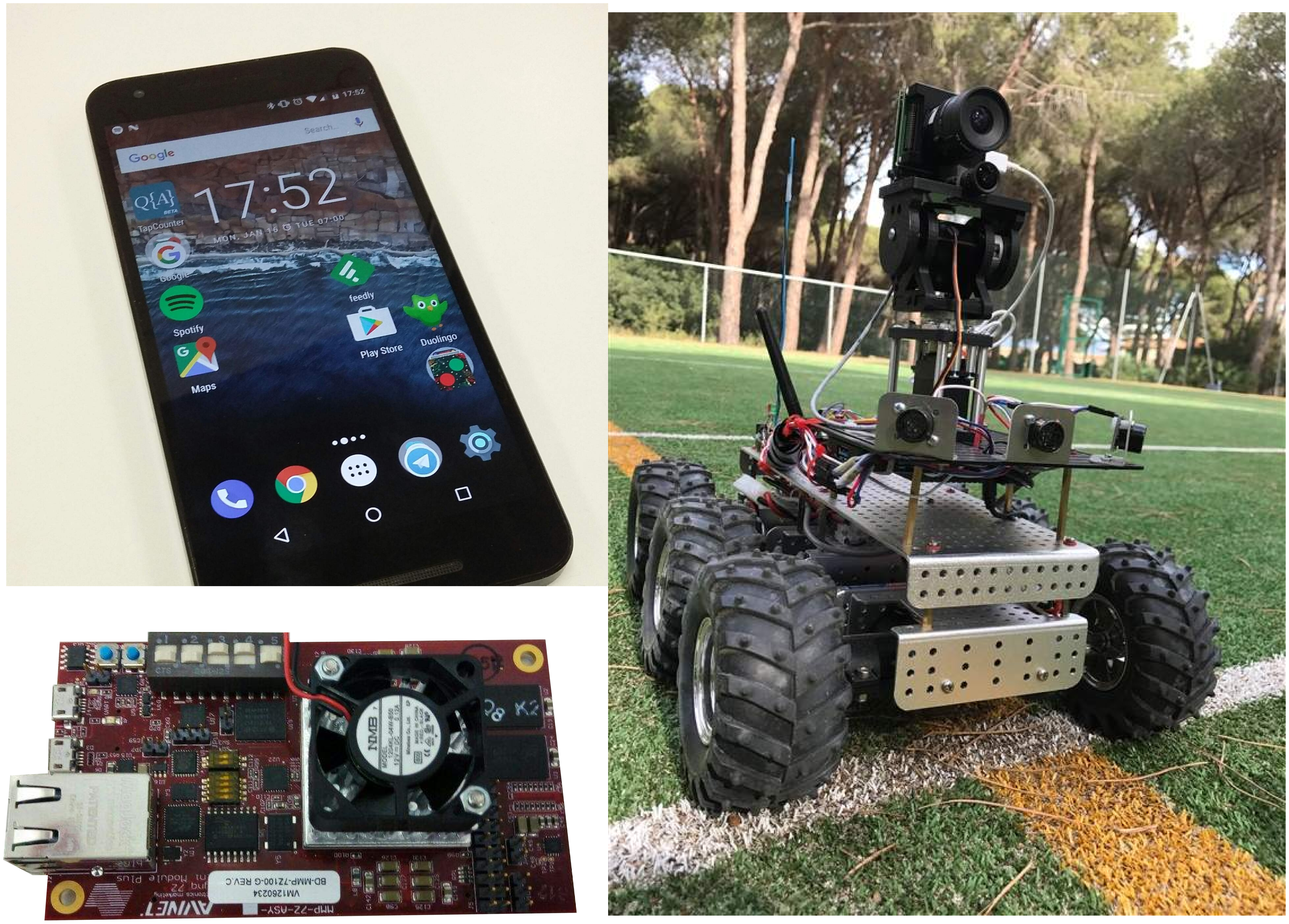}
        \caption{}
    \end{subfigure}
    \caption{Motivation for reduced precision in weights and activations of deep \aclp{cnn}. (a) Memory arrangement and comparison in terms of energy consumption, latency and capacity. (b) Target platforms for convolutional neural networks.}
    \label{fig:memory_motivation}
\end{figure*}

In the last decade, machine learning applications based on \acf{cnn} have gained substantial attention due their high performance on classification and localization tasks~\cite{lecun_deep_2015}. 
In parallel, dedicated hardware accelerators have been proposed to speed up inference of such networks after training is completed \cite{Aimar_etal17, Chen2016, Han,Conti,Jouppi2017}. 
Since the target of such hardware accelerators are mobile devices, IoT devices and robots, reducing their memory consumption, memory access and computation time are crucial (see Fig.~\ref{fig:memory_motivation})
Pruning and model compression \cite{Han_etal15}, quantization methods \cite{Courbariaux_etal14, Courbariaux_etal15,Mueller_Indiveri15, Gupta_etal15,Hubara_etal16}, as well as toolboxes \cite{Gysel_etal16, Tensorflow15,pytorch} have been developed to reduce the numbers of neurons and connections.
The deployment of \acp{cnn} on embedded platforms with only fixed point computation capabilities and the development of dedicated hardware accelerators has also spurred the development of toolboxes that can train CNNs for fixed point representations \cite{Tensorflow15,pytorch,Gysel_etal16}.
The popular toolbox Ristretto ~\cite{Gysel_etal16}, for example, can be used to train \acp{cnn} with fixed-point weights, but not fixed-point activations.
To adapt both weights and activations of \acp{cnn} trained on conventional GPUs using 32 bit floating-point representation to fixed-point hardware, we developed the ADaPTION toolbox\footnote{Code available: https://github.com/NeuromorphicProcessorProject/ADaPTION} (see Fig.~\ref{fig:motivation}). 
Furthermore, our toolkit supports also to train \acp{cnn} from scratch with specified precision for both weights and activations to run on the recently developed hardware accelerator NullHop \cite{Aimar_etal17}.\\
In the next section we will introduce new functionalities and parameters added to Caffe, as well as the workflow of ADaPTION. 
In Section \ref{sec:discussion} we will discuss the crucial components to achieve SOA classification accuracy with reduced precision weights and activations and compare ADaPTION directly to other existing toolboxes.\\

\section{Low-precision add-on for Caffe}
\begin{figure}[!t]
\centering
\includegraphics[width=\textwidth]{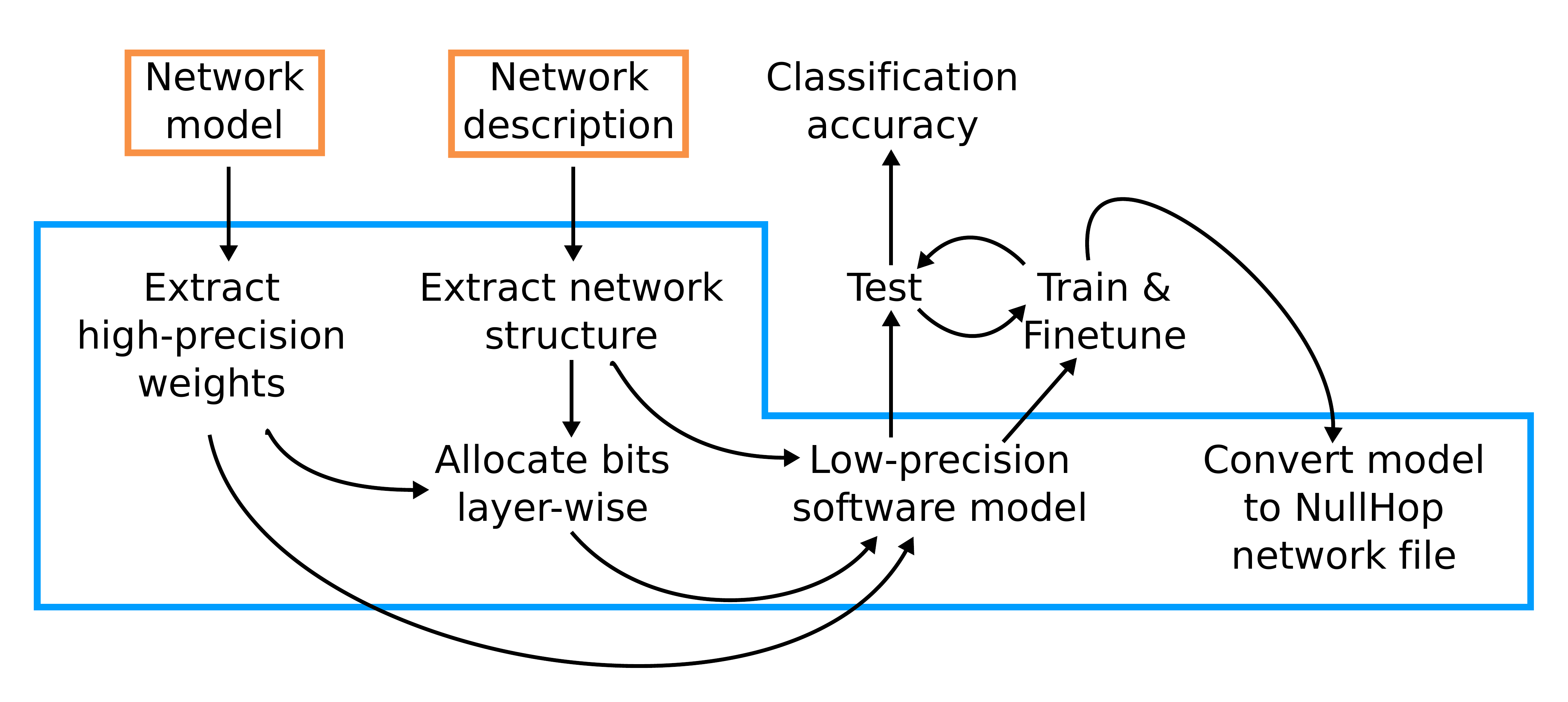}
\caption{Pipeline overview: ADaPTION extracts the network structure and the pre-trained weights. The total number of available bits is distributed for weights and activations independently, as well as independently for each layer to create a low-precision model. The model is then fine-tuned or trained from scratch to solve the desired task. Once acceptable classification accuracy is reached the low-precision model can be converted to Nullhop compatible file format.}
\label{fig:pipeline}
\end{figure}

\noindent Caffe is a deep learning library developed by the Berkeley Vision and Learning Center \cite{Jia_etal14}. 
It provides all state-of-the-art error backpropagation gradient descent tools such as ADAM, ADAGrad and many different layer types, as well network architectures. 
To incorporate low-precision training within this framework we added three new layer types: LPInnerProduct, LPConvolution and LPAct, where the LP stands for low-precision. 
These layers operate the same way as their high-precision counterpart, except that during the forward pass the values are quantized to the respective fixed-point representation given a specified bit-precision and decimal point, e.g. signed 16 bit, Q1.15\footnote{The notation Qm.f sets the decimal point of a fixed-point number, where m represents the integer part and f the fractional part of the number.}. 
To round the weights and activations we introduced three additional parameters: \textsf{BD}, \textsf{AD} and \textsf{rounding\_scheme}, which are specified in the network configuration file. 
The parameter \textbf{BD} (\textbf{B}efore \textbf{D}ecimal point) specifies the \textit{maximum integer value} that can be represented. 
It determines the number of bits of the integer part of the respective value is allowed to occupy, including the sign bit.
The parameter \textbf{AD} (\textbf{A}fter \textbf{D}ecimal point) specifies the \textit{precision} that can be represented. 
It determines the number of bits the fractional part of the respective value is allowed to occupy.
The rounding\_scheme parameter is a flag that sets the the option to either round weights and activations deterministically or stochastically.
The notation Qm.f is implemented in Caffe as QBD.AD.\\\\\\
Our toolkit ADaPTION has the following features (see also Fig. \ref{fig:pipeline}):
\begin{itemize}
\item extraction of network structure into a \textit{net\_descriptor}
\item dynamic, layer-wise distribution of predefined available number of bits for weights and activations independently according to the respective dynamic range
\item creation of a new low-precision network based on the extracted or user-defined net\_descriptor, as well as layer-wise or global bit-distribution
\item fine-tuning\footnote{A network subject to fine-tuning is initialized with pre-trained weights, which are subject to further training on the same data set.} of extracted high-precision weights or re-training from scratch with user-specified rounding\_scheme
\item exporting the network to a NullHop \cite{Aimar_etal17} compatible file format
\end{itemize}
\noindent These changes are fully compatible with the original version of Caffe and can be merged or used as a stand-alone Caffe version.\\

\subsection{ADaPTION workflow and method}
For adapting the quantized weights and activations, we used the method called \textit{power2quant}, formerly known as dual copy rounding, developed by \cite{Stromatias_etal15} and concurrently by \cite{Courbariaux_etal14}. ADaPTION works in the following way: it extracts the network structure and the pre-trained weights of a given \ac{cnn} model. 
The structure is adapted to use low-precision convolutional, \ac{relu} and fully-connected layers provided as separated layers. 
The low-precision activation layer is separate from the actual activation layer in Caffe. 
The activations are quantized before they are sent to the activation layer, i.e. the \ac{relu} layer. 
The separation of quantization procedure and activation function  has the potential advantage that new activation functions can be directly used in ADaPTION.
The pre-trained weights are converted into the low-precision Caffe blob\footnote{A blob is a data storage structure that is used by Caffe to store the neuron parameters, such as the weights or biases.} structure.  
An ADaPTION method measures the dynamic range of weights and activations by inferring a random set of training images. 
The measured dynamic ranges are used by another method to iteratively allocate the total number of bits (specified by the user according to their needs) between the integer and fractional part of the fixed-point representation, as explained in Sec.~\ref{sec:layerwise}. 
The low-precision blob structure, as well as the layer-wise bit distribution is then used to generate the low-precision model. The model can either be initialized using random weights or using the pre-trained high-precision weights. The latter normally results in faster convergence, as well as higher classification accuracy.
In the beginning we allocate for each layer two weight and two bias blobs.
One blob is used to perform inference, which will quantize the weights, biases and activations to its specified bit precision. 
The second blob is used to calculate the gradients during training. 
Once the classification accuracy is close to floating-point network level, or it does not change anymore we stop the training. 
The resulting Caffe model can then be converted to a specific hardware accelerator format, such as NullHop \cite{Aimar_etal17}.\\

\begin{figure}[!t]
\centering
\includegraphics[width=\textwidth]{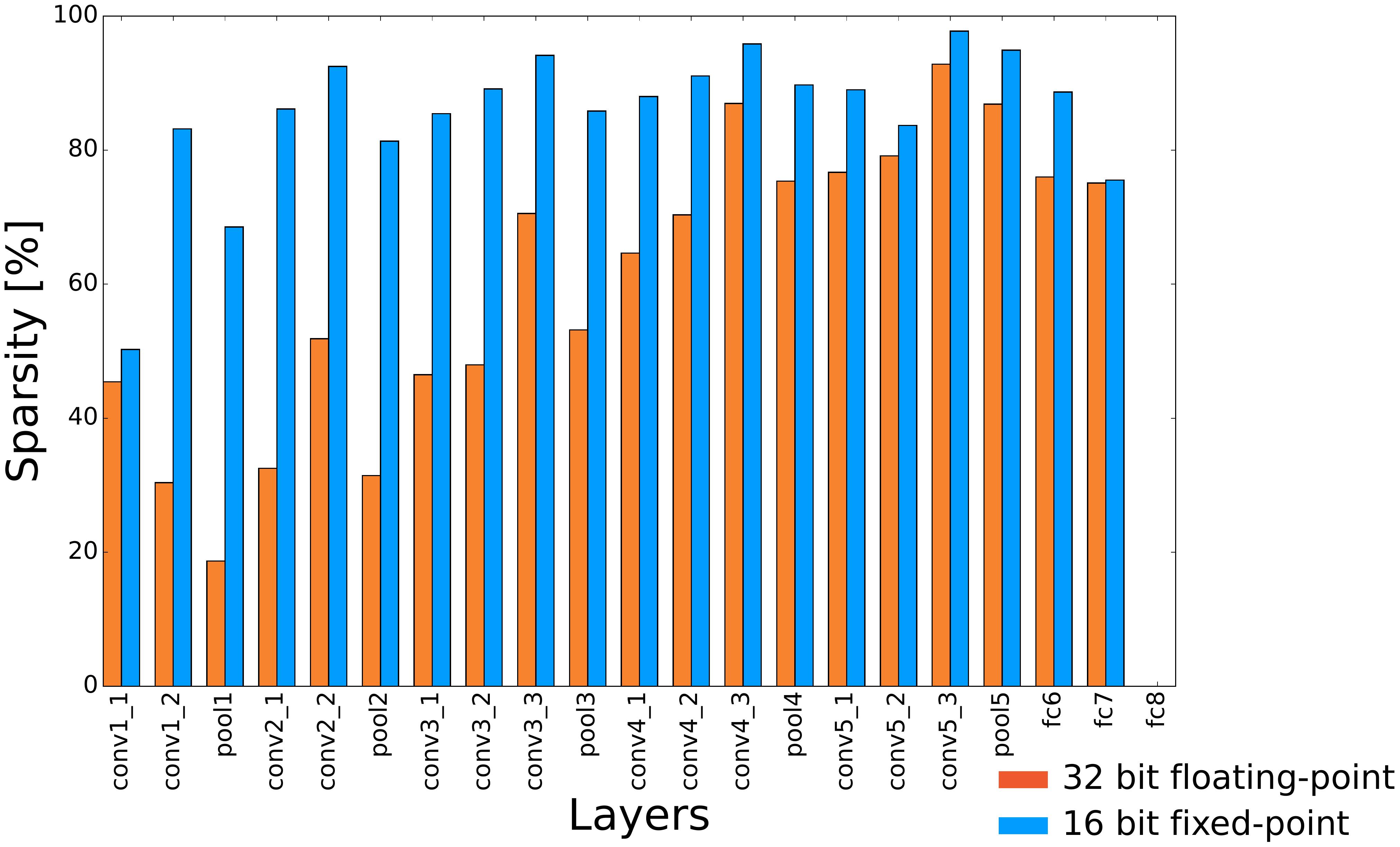}
\caption{Comparison of sparsity in activations of VGG16: Sparsity within each layer of VGG16 classifying 25,000 random test images from ImageNet. The high sparsity (up to $\sim$50\% increase) can only be achieved if network is fine-tuned to a given fixed point bit-precision.}
\label{fig:sparsity}
\end{figure}

\subsection{Effect of one-time rounding on sparsity \& classification accuracy}
Training large networks, such as VGG16, is time consuming.
Thus, we investigated if it is possible to perform quantization in a single step, without fine-tuning or training from scratch.
One-time rounding of weights to 16 bits with reasonable decimal point does not impair classification accuracy, if  the activations are kept at 32-bit floating-point. 
If one quantizes the weights down to 16-bit fixed-point, it turns out that $\sim$ 90\% of them can be represented with only 4 bits and $\sim$99\% with 8 bits.
Even reducing the maximum number of available bits for weights down to 8 bits does not severely affect classification accuracy ($\sim$ 65\%). This finding clearly shows that the weights are not the limiting factor when a full precision network is quantized to reduced precision

On the other hand, one time-rounding (deterministic or stochastic; see Sec.~\ref{sec:rounding}) of weights, and reducing activations 16-bit fixed point without fine-tuning of weights reduces accuracy to chance level.
The level of sparsity is also increased much less than if we fine-tune the network.
These results suggest that quantizing the activations is actually a difficult problem, 
since the network's performance depends more on the available dynamic range of activations than on the weights.

The average sparsity in activations of VGG16 is 57\% after quantizing weights and activation down to 16 bit without fine-tuning, however the classification accuracy drops to chance level. 
In contrast, if we fine-tune the network with quantized weights and activations to a single global fixed point representation (e.g. Q8.8), we achieve an average sparsity of 82\% and a Top-1 classification accuracy of 59.4\%. 
Sparsity values are obtained using 25 000 random images from the test set (Fig. \ref{fig:sparsity}).
The increase in sparsity of activations, especially in early layers of VGG16, of up to ~50\% can only be achieved if the network is trained from scratch or if it is fine-tuned using reduced precision weights and activations. 
This sparsity can optimally be exploited by the NullHop hardware accelerator to efficiently skip computations \cite{Aimar_etal17}.

\subsection{Layer-wise quantization}
\label{sec:layerwise}
\begin{figure}[!h]
\centering
\includegraphics[width=\textwidth]{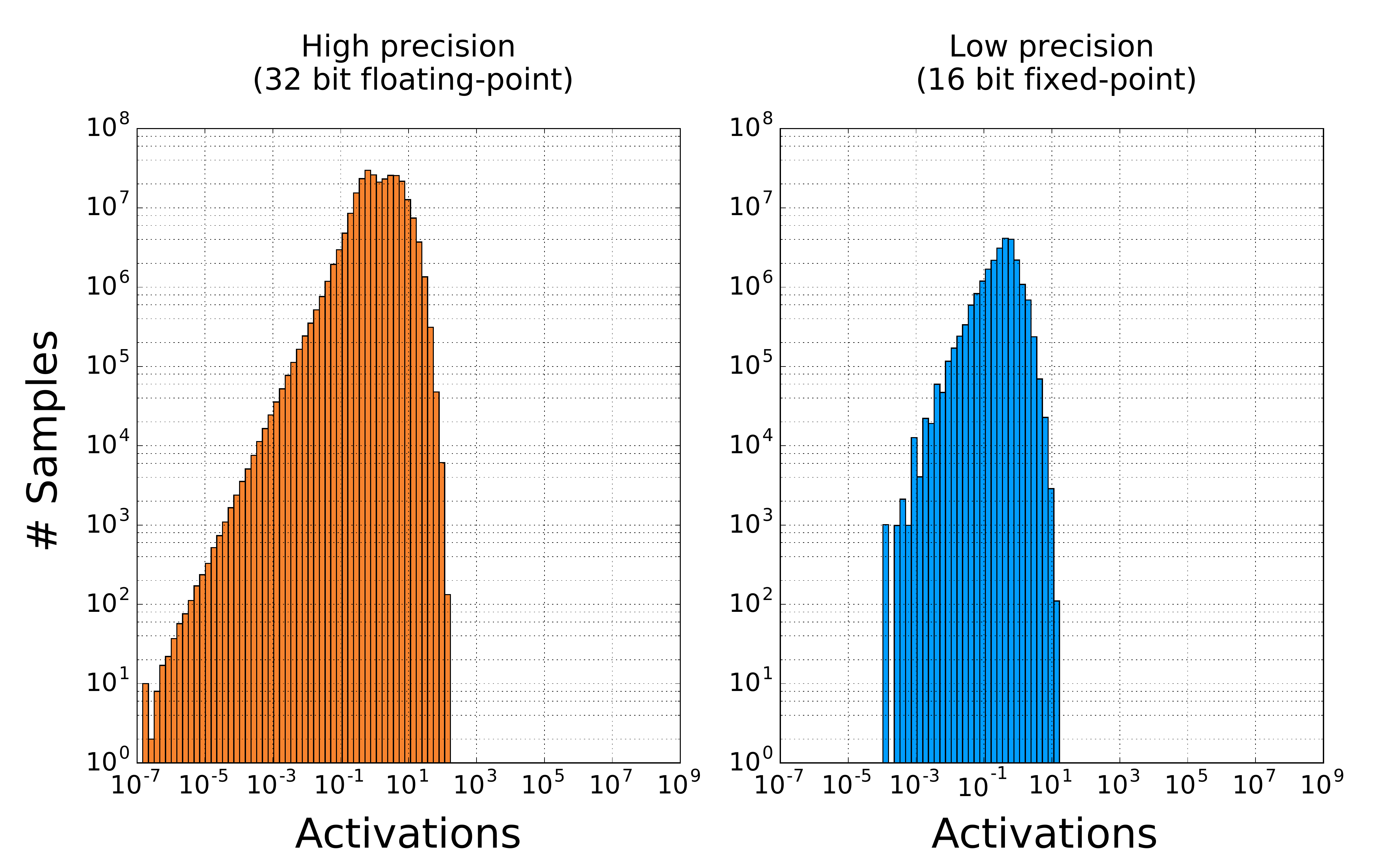}
\caption{Dynamic range of activations in high (orange) and low (blue) precision setting after training. Quantization acts as regularizer to keep the activations in a resolvable range and also prevents the activations from saturating.}
\label{fig:act_comp}
\end{figure}

The activations in the original VGG16 typically span 9 orders of magnitude from $10^{-4}$ up to $10^5$ (see Fig.~\ref{fig:act_comp}), whereas the dynamic range of the activations, achieved with training the network with reduced precision spans only 4 orders of magnitude ($10^{-3}$ to $10^1$). 
Thus, constraining the weights and activations to a given fixed-point representation has the effect that the dynamic range requirement is reduced by 5 orders of magnitude.
However, even if we choose two different decimal point locations for weights and activations, i.e. Q2.14 for weights and Q14.2 for the activations, globally for all layers, the Top-1 classification accuracy drops to 8.7\% after quantizing the network once. 
After fine-tuning, we could achieve 59.4\% Top-1 classification accuracy, which is merely ~9\% below its high-precision counterpart. 
The observed drop in classification accuracy suggests that the original VGG16 needs the entire dynamic range in each layer of its activation and is not able to produce state-of-the-art classification accuracy if the activations are bounded by a single given fixed-point representation.
We investigated the dynamic range for each layer separately. 
We found that the layer-by-layer dynamic range especially of the activations differs significantly. 
Therefore, we implemented a per-layer decimal point.
A similar approach to compress the activation functions precision per-layer, has been proposed earlier \cite{Wang_etal16}.
However, while \cite{Wang_etal16} perform rigorous mathematical assumptions on how to distribute the available number of bits for a given layer, we propose a simple, computationally cheap iterative scheme.
In our method, we keep the overall number of available bits fixed and check iteratively how many weights/activations cannot be represented if we reduce the available number of bits for the integer part. 
In a second step we check if the percentage of lost weights/activations is below a user defined threshold (usually below 1\%). 
We are looking at the integer part of the weight/activation, since we can check against the maximum number present in a given layer. 
If we would look at the smallest number we could not directly link this value to the required number of bits, since we can not easily access the precision needed to represent small values. 
By locating the decimal point individually for each layer according to our proposed scheme, we achieve a Top-1 classification accuracy of 64.5\%.
Even though  in this study we used 16-bits with per-layer decimal point, ADaPTION can be used to choose any precision and any decimal point for any layer separately, .

\subsection{Deterministic vs. Stochastic rounding}
\label{sec:rounding}
With the aforementioned additions to ADaPTION we were able to increase the Top-1 classification accuracy from 59.4\% up to 64.5\% by fine-tuning the network using deterministic rounding. 
However, training with reduced precision and the dual-copy scheme introduces inescapable fixed points in parameters space (see Sec. \ref{sec:insights} for more details).\\
To counteract these fixed points and to increase classification accuracy we investigated the effect of using stochasticity during training. 
Stochastic rounding, in contrast to deterministic, has the advantage that each step during training has pseudo-randomness. Stochastically rounded values have the correct expectation value but are drawn from a binomial distribution of the two nearest fixed point values.
Using any kind of stochasticity usually yields networks that have better generalization properties, tend to have higher classification accuracies, and prevent overfitting~\cite{Srivastava_etal14}. 
In the context of low-precision quantization, stochastic rounding not only provides higher classification accuracies, but it is helpful, if not crucial, to successfully avoid the inescapable local fixed points introduced by the low-precision training.
The training convergence time stayed the same as in the deterministic case.

Using the same number of training examples, we finally achieved a Top-1 classification accuracy of 67.5\%,  which is only 0.8\% below its full-precision counterpart.
We added the option to ADaPTION to control the rounding scheme, and added the necessary function in the quantization algorithm. 
To use stochastic rounding for a given layer the \texttt{rounding\_scheme} option must be set to \textit{STOACHASTIC}.

\subsection{Training VGG16 with reduced-precision on weights and activations}
\paragraph{Insights}
\label{sec:insights}
Like in other studies~\cite{Courbariaux_etal16}, we found that hyperparameters such as the learning rate must be 10 to 100 times smaller during training compared to high-precision network to ensure convergence: Smaller jumps further ensure that fixed points in parameter space caused by quantization are avoided. 
These fixed points represent parameter combinations that lead to sudden tremendous decreases in accuracy. For example, we observed single-iteration jumps from 59\% accuracy down to chance level. After these large jumps, the network has to start training again from scratch.
The lower learning rate results in longer training time \cite{Mishra_etal17}, especially when training low precision networks and also using low-precision gradients \cite{Courbariaux_etal14, Courbariaux_etal16, Zhou_etal16}.\\
During training, the gradient is calculated based on the full precision 32-bit floating-point weights and activations, that are only quantized during inference, i.e. per batch.
This training scheme is different from low-precision training in which the gradient is calculated based on the quantized weights and activations and the gradient itself is constrained to a low-precision fixed-point number. 
Low- or even extreme low-precision training has been investigated by \cite{Rastegari_etal16,Zhou_etal16}, but so far the resulting accuracies are not competitive.
Furthermore, weights must be initialized taking the respective fan-in and fan-out of each unit into consideration \cite{Glorot_Bengio10}. 
In order to keep the activation introduced by the stimulus (image) in a range that can be resolved by the first low-precision convolution layer, we use scaling. The scaling parameter normalizes pixel intensities so that the highest value does not saturate the integer bit range precision. 
Scaling uses a single scalar value that multiplies each pixel value.
Without scaling the input, the activations saturate at the maximal possible value and the images are harder to classify. 
Hence, with scaling the full dynamic range of possible values can be used, which speeds up training and leads to higher classification accuracies. 
These insights and factors are necessary to reach state-of-the-art classification accuracy.

Stochastic rounding enabled the low-precision network to reach state-of-the-art classification accuracy (Top-1 67.5\% vs 68.3\% in high-precision network \cite{Simonyan_Zisserman14}).


\section{Discussion}
\label{sec:discussion}
\subsection{Benefits \& Limitations of ADaPTION}
The direct effect of quantizing the activation to an intermediate precision, such as 16 bit fixed-point, is that the average sparsity. i.e. the number of zero elements divided by the total number of elements, is at 82\%, whereas in the high-precision case the average sparsity is only 57\%. 
Especially early and early-intermediate layers show a much higher sparsity when quantized.
In these layers the sparsity increased by up to 50\%, while the average increase was 25\%.
This suggest that features in early-intermediate layers are not as crucial as late layers to perform a correct image classification.
This sparsity can be exploited, for a hardware accelerator that skips multiplication if the includes zeros. 
The NullHop accelerator is an example \cite{Aimar_etal17}.\\
The secondary effect of quantization is that it acts as a regularizer to keep the activations and the weights in resolvable range, without saturation effects (Fig. \ref{fig:act_comp}).\\
Sparsity in weights is not as strongly affected as activations. 
The smaller weight sparsity is because weights tend to be quite small and centered around zero in 32 bit floating-point. 
Thus, weight quantization has only an effect if the available number of bits drops below 8 bits. 

ADaPTION supports any bit-distribution down to a single bit.
However, detailed analysis of extreme low-precision quantization is beyond the scope of this paper and has been analyzed in \cite{Courbariaux_etal15,Lin_etal15,Rastegari_etal16, Mishra_etal17} 
We can propose a close-to-optimal decimal point location per layer, if we can check the dynamic range, using for example a pre-trained high-precision model. 
Without this check, for example with a complete new architecture directly trained with reduced precision, it is not a straightforward process to allocate the available number of bits.
One way to allocate the available number of bits is to set a fixed-global distribution, e.g. Q8.8 for weights and activations. For smaller networks this was sufficient \cite{Lungu_etal17,FaceNet}. 
Another way of doing this is to first train a model without quantizing the weights and activations and fine-tune it afterwards to the desired bit precision. 
It has the drawback of long training time and high computational cost.\\
A trend we observe is that weights tend to be very small and always centered around zero. 
Reserving just 2 bits (including the sign) for the integer part is usually enough. 
Activations, in contrast, tend to be quite large in early layers, thus requiring more bits for the integer part compared to later layers.
Activations decrease significantly with depth in the network and thus require more bits for the fractional part.\\
\noindent As direct consequence of reduced bit precision, we achieve a reduction of the overall memory consumption: this reduced memory footprint can improve the performance of generic hardware but we expect it to be particularly significant for custom hardware architectures.
Similarly, quantization has the useful side effect that sparsity, i.e. the percentage of zero-valued elements, is increased, which can be exploited by hardware supporting sparse convolution operations to save power and reduce the computation time.

\subsection{ADaPTION vs. Ristretto}
During the development of ADaPTION, Gysel and colleagues developed an add-on for Caffe called Ristretto \cite{Gysel_etal16}. 
Ristretto positions itself in a similar role as ADaPTION, but lacks key-features which makes
quantization quite difficult and at the same time interesting for hardware accelerators, i.e. fixed-point activations to skip multiplication with zero. 
Furthermore, this key feature is crucial for our Nullhop. 
Ristretto does not support a pre-defined number of bits to distribute between integer and fractional part of a fixed-point number. 
Furthermore, the number of bits, which is provided by Ristretto, does not stay constant across the network, which is crucial for hardware accelerators, such as Nullhop. 
Probably most important, Ristretto also does not support fixed-point rounding of activations, which, as we showed, contributes the most to the sparsity in activations and is the hardest to constrain to a given bit-precision and still provide state-of-the-art classification accuracy. 

\section{Conclusion}
We presented ADaPTION, a new toolbox to quantize existing high-precision \acp{cnn}s to be efficiently implemented on dedicated mobile-orientated hardware accelerators. The toolbox adapts weights and activations to globally fixed or layer-wise fixed-point notation.\\
Quantization of weights and activations has the advantage that the overall sparsity in the network increases while preserving state of the art Top-1 ImageNet classification accuracy.

\section*{New Benchmark Networks}
A major problem while comparing \ac{cnn} accelerators is that many works use custom networks, making the comparison between different architectures hard.  
Even if some architectures (e.g. VGG16, GoogLeNet, ResNet) are more popular than others, they cover a limited range of hyperparameters and computational costs and are not ideally-suited for realistic hardware benchmarking. 
In order to address this issue we used our software framework to train a new \ac{cnn} architecture characterized by a variety of kernel sizes and number of channels, useful to verify hardware computational capabilities in multiple scenarios. 
Due to the fast-changing \acp{cnn} and other DNN accelerators landscape, we will update this paper, adding new networks suited for benchmarking new emerging hardware designs.

\subsection*{Giga1Net}
\textsf{Giga1Net}, defined in Table~\ref{tab:giga1net_specs}, requires 1~GOp/frame to classify an image from the ImageNet dataset. The prototxt necessary for the training is available in the ADaPTION repository; Giga1Net achieves 38\% ImageNet Top-1 accuracy after 36h of training on a GTX980 Ti.

\begin{table}
\centering
\caption{Giga1Net inference parameters}
\label{tab:giga1net_specs}
\begin{tabular}{|l|c|c|c|c|c|c|c|}
\hline
\begin{tabular}[c]{@{}l@{}}Layer\end{tabular}                                                                
& \begin{tabular}[c]{@{}l@{}} Input\\feature\\maps\end{tabular}                                                    
& \begin{tabular}[c]{@{}l@{}} Output\\feature\\maps\end{tabular} 
& \begin{tabular}[c]{@{}l@{}} Kernel\\Size \end{tabular}                  
& \begin{tabular}[c]{@{}l@{}} Input\\Width/Height\end{tabular}                                                     
& Pooling 
& \begin{tabular}[c]{@{}l@{}}ReLU\end{tabular}      
& Stride
\\ \hline
1 - conv& 3 & 16 & 1 & 224x224 & Yes & Yes & 1\\
2 - conv& 16 & 16 & 7 & 112x112 & Yes & Yes  & 1\\
3 - conv& 16 & 32 & 7 & 54x54 & Yes & Yes & 1\\
4 - conv& 32 & 64 & 5 & 24x24 & No & Yes & 1\\
5 - conv& 64 & 64 & 5 & 22x22 & No & Yes & 1\\
6 - conv& 64 & 64 & 5 & 20x20 & No & Yes & 1\\
7 - conv& 64 & 128 & 3 & 18x18 & No & Yes & 1\\
8 - conv& 128 & 128 & 3 & 18x18 & No & Yes & 1\\
9 - conv& 128 & 128 & 3 & 18x18 & No & Yes & 1\\
10 - conv& 128 & 128 & 3 & 18x18 & No & Yes & 1\\
11 - conv& 128 & 128 & 3 & 18x18 & Yes & Yes & 1\\
12 - FC& 128 & 4096 & -  & - & No & Yes & -\\
13 - FC& 4096 & 1000 & - & - & - & - & -\\
\hline
\end{tabular}
\end{table}
\section*{Acknowledgements}
This work was funded in part by Samsung. The authors thank Matthieu Courbariaux for helpful comments on the manuscript and Lorenz Mueller for his critical comments on the toolbox and interpretation of the results. Further, we thank Iulia Lungu and Enrico Calabrese for testing the library and providing more feedback. Also thanks to Federico Corradi for incorporating ADaPTION in the cAER library. 

\newpage
\bibliographystyle{plain}

\end{document}